\documentclass[letterpaper, 10 pt, conference]{ieeeconf}  

\IEEEoverridecommandlockouts                              

\usepackage{graphicx}
\usepackage{kotex}
\usepackage{tabularx}
\usepackage{adjustbox}
\usepackage{url}
\usepackage{color}
\usepackage{cite}
\usepackage{array}
\usepackage[table]{xcolor}
\usepackage{booktabs}
\usepackage{bbding}
\usepackage{pifont}
\usepackage{tabularray}
\usepackage{colortbl}
\usepackage{amsmath,amssymb}
\usepackage{caption}
\usepackage{multirow}
\usepackage{nicematrix}
\usepackage{float}
\title{\LARGE \bf
Which Terrain Is Better? Preference Learning with VLM Prototypes for Off-Road Traversability Ranking}

\author{Ji-Hoon Hwang, Jisung Bae, E-In Son, Dong-Wook Kim, Jung-Taak Kim and Seung-Woo Seo}

\begin{document}
\captionsetup[table]{position=bottom}
\captionsetup[figure]{skip=3pt}

\maketitle
\thispagestyle{empty}
\pagestyle{empty}

\begin{abstract}
In vision-based off-road navigation, a robot needs to know not only which obstacles to avoid but also which terrain is better.
The first is handled by freespace detection or semantic segmentation. The second is usually answered with a traversability score, but no universal ground truth exists for such a score, so perception falls back on a predefined value per semantic class or a freespace confidence.
These scores say what a region is, not which region a robot should prefer.
We therefore formulate this preference as visual traversability ranking, an ordering of visible terrain that can be supervised by comparisons between two regions.
Standard annotations do not label preference, but they imply its direction.
We present TravPro, which converts these annotations into ordered region pairs and fits a small readout on frozen vision--language model (VLM) patch tokens to these pairs.
The tokens are clustered once into a fixed prototype bank, and the readout learns a preference score per prototype.
The readout is then applied to every patch and serves as a teacher that turns sparse comparisons into dense preference pseudo-labels without pixel-wise annotation.
An RGB student distills these maps into a dense terrain-preference map together with a non-ground mask that excludes obstacles and background from the ranking.
On five unseen domains, TravPro reaches a mean pairwise accuracy of 0.915 against 0.783 for the strongest baseline, producing an ordering sensitive to surface condition that a per-class value cannot represent.
The same VLM and the same supervision yield no such ordering when the VLM is prompted and the supervision is used as dense targets; what matters is how they are used.
\end{abstract}

 
\section{INTRODUCTION} 
\label{sec:intro} 
\begin{figure}[t!]
\centering
\includegraphics[width=0.97\linewidth]{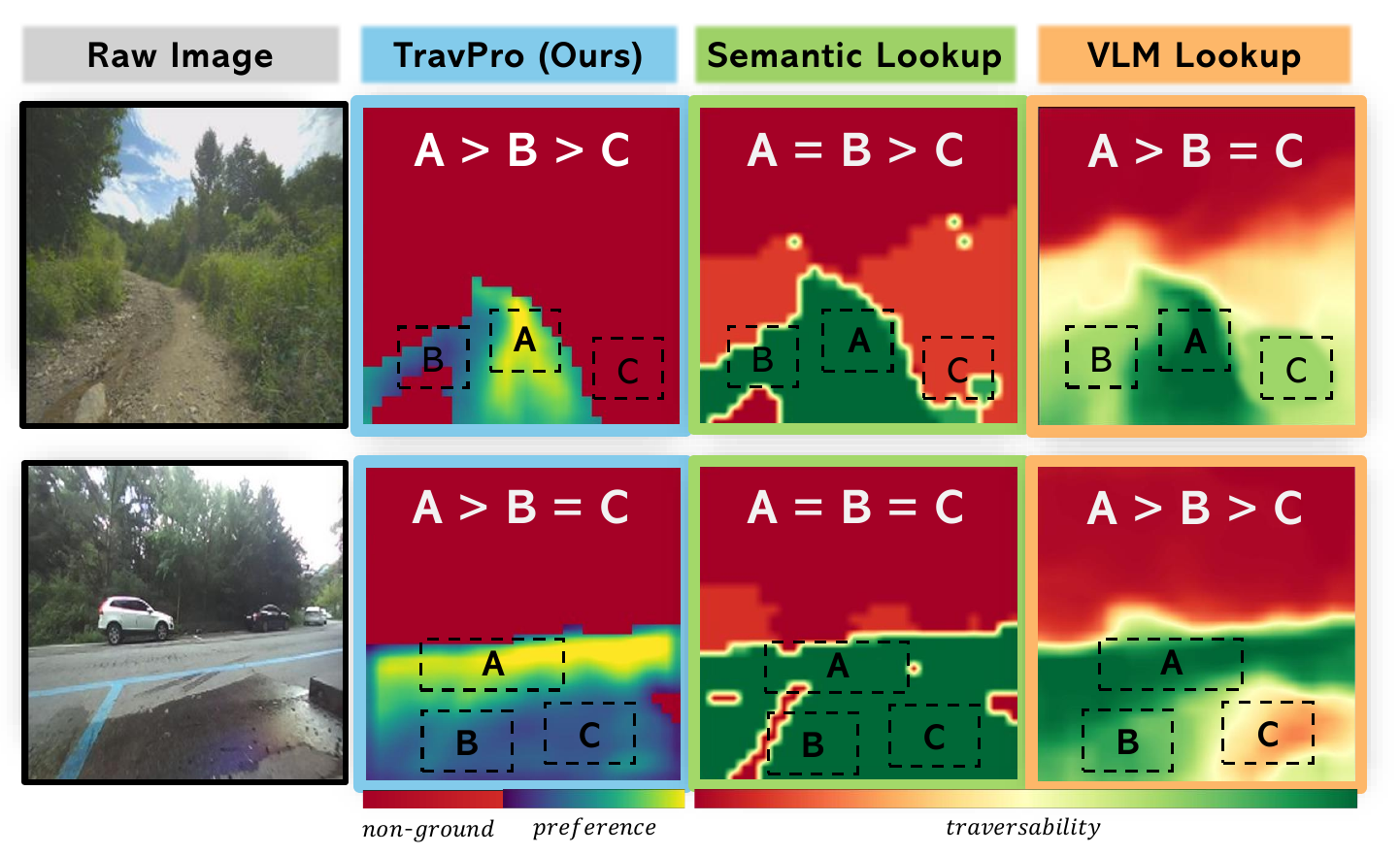}
\captionsetup{font=small}
\caption{Semantic and VLM lookups fail to capture terrain preference. Two scenes, each with three marked regions, and how each method orders them. A semantic traversability lookup gives regions that share a label the same fixed value, since its taxonomy does not define surface condition. A VLM lookup does the same with classes prompted from an open vocabulary, so the vocabulary is wider but the order is still fixed per class. TravPro (ours) learns the preference from ordered region pairs.}
\vspace{-0.5cm}
\label{fig:teaser}
\end{figure}

Autonomous navigation in unstructured outdoor environments requires two decisions from perception: which regions to avoid, and which of the remaining ones to prefer.
Vision-based off-road perception addresses the first decision through freespace detection or semantic segmentation~\cite{orad,rugd,rellis,goose}.
The second is answered only as a by-product: a fixed value per semantic class, or the confidence of a freespace detector.
Neither is trained to reflect preference.
Fig.~\ref{fig:teaser} shows the consequence, that a model trained to recognize soil gives every soil region the same value.
We formulate the second decision as \emph{visual traversability ranking}, an ordinal preference over visible terrain, since no ground truth exists for an absolute traversability score.

The two decisions are not supervised equally.
The first has dense labels, since freespace and semantic annotations both mark what is not ground.
The second has no labels, but the annotations still imply a direction of preference.
Semantic categories give relations such as asphalt over gravel, and some freespace annotations divide the drivable region into a few classes that can be related in the same way.
Surface condition, what weather or damage have done to a material, appears in these annotations only through a few labels such as puddle and water; they are too rare to learn as dense targets, but as one side of a relation such as soil over puddle they give a reliable direction.
RSCD~\cite{rscd} adds many more such relations, wet, rough, and snow-covered surfaces of the same material, one label per close-up road-surface image.
The viewpoint differs from a robot's, but a dry and a wet surface of the same material stay ordered regardless.
These relations differ in scale, from region to whole image, but each says the same thing, that one region is preferred to another.
The problem is to assemble them into one dense field that distinguishes surface conditions and transfers to domains that provide no such relations.

Existing approaches obtain preference in three ways.
A semantic traversability lookup~\cite{stseg} attaches one fixed score to each category, so it can express a condition only where the taxonomy draws a class boundary.
Sparse pairwise methods such as W-RIZZ~\cite{wrizz} learn from manually annotated point pairs, which removes the need for dense labels but requires new pairs for every domain, since what is learned from one set does not transfer to another.
Embodied and self-supervised methods derive preference from demonstrations, trajectories, or proprioception~\cite{wvn,vstrong,gronav,hindel}, so the resulting order reflects one platform in one environment and extending it requires further traversal.
In every case, preference is either fixed in advance for each class or collected again at the site where the model is used. None of them learns it from annotations that already exist.

Vision--language models (VLMs), trained on web-scale image--text data, carry a general recognition of what surfaces are and how they look, and may fill this gap.
Recent work taps this recognition directly, asking the VLM for a traversability score or prompting an open-vocabulary segmenter with a fixed set of terrain classes weighted by predefined values (Fig.~\ref{fig:teaser})~\cite{anytraverse,zest,catnav}.
In both, the score is not fit to region comparisons.
A prompted score changes with the prompt and the image, and a weighted class score is fixed per class.
The problem is in the readout, not in the features.
We hypothesize that the patch tokens of a frozen VLM already separate terrain categories as well as surface conditions; what they lack is an ordering, which can be learned on top without changing them.

We present TravPro, which learns this ordering on top of a frozen VLM from the relations that the annotations already imply, using the VLM as a training-time teacher rather than an inference-time scorer.
The non-ground labels supervise the first decision directly; the remaining annotations are converted into ordered region pairs, each giving a direction of preference and no magnitude.
Frozen VLM patch tokens from off-road scenes and close-up road-surface images are clustered once into a fixed bank of prototypes.
Since the bank is built from appearance and not from category, two surfaces of the same material in different conditions can land on different prototypes.
A readout on that bank is then fit from the pairs, one preference score per prototype.
Applying the same readout to every patch turns sparse region-level and image-level comparisons into a dense pseudo-label map.
An RGB student distills this map and, together with the non-ground labels, learns both decisions. At test time it takes RGB alone and outputs a terrain preference and a non-ground mask in a single forward pass.
Optionally, a small residual head adapts the ranking to a new site from a few target pairs while the student stays frozen, since preference in the field varies with site, platform, and driving policy.

Our main contributions are as follows:
\begin{itemize}
\item We formulate visual traversability ranking as an ordinal preference over visible terrain, and show that existing freespace, semantic, and road-surface annotations together supply enough ordered pairs to learn it.
\item A prototype preference teacher built on frozen VLM tokens that turns these pairs into dense preference targets, distilled into an RGB student that outputs terrain preference and a non-ground mask from one forward pass.
\item Results on five unseen domains, where TravPro reaches a mean pairwise accuracy of 0.915 against 0.783 for the strongest baseline, 0.581 for the same VLM prompted directly, and 0.723 for a lookup trained on the same supervision as dense targets.
\end{itemize}

\section{RELATED WORK}
\label{sec:related}

\subsection{Off-Road Perception and Traversability}
Vision-based off-road perception has been studied mainly as freespace detection and semantic segmentation.
ORAD-3D divides drivable ground into a few classes and marks the rest as non-drivable~\cite{orad}, and RUGD, RELLIS-3D, and GOOSE provide dense semantic labels for terrain, vegetation, objects, and structures~\cite{rugd,rellis,goose}.
Neither label type provides a preference order within the drivable terrain.
A semantic traversability lookup assigns a fixed score to each category and therefore cannot represent a change of condition within a category.

W-RIZZ~\cite{wrizz} is the closest to our formulation.
It learns a continuous traversability field from sparse ordinal point pairs, does not assume constant traversability within a class, and labels pairs across images so that predictions stay consistent between images.
Its pairs are annotated by hand on the target dataset, and it also shows that pairs can be generated from the traversability tiers of a semantic taxonomy.
Embodied and self-supervised methods instead derive traversability from demonstrations, traversed trajectories, or proprioceptive feedback~\cite{wvn,vstrong,gronav,hindel}, which ties the result to the platform and environment where the data were collected.
TravPro differs in how it obtains and uses pairs.
The pairs come from existing annotations, including image-level condition labels that no segmentation taxonomy expresses.
A readout on frozen VLM prototypes is fit to these pairs and turns them into dense targets for an RGB student.

\subsection{Foundation Models for Traversability and Distillation}
Foundation models have recently been applied to traversability.
ViTA~\cite{vita} adapts SAM2~\cite{sam2} with risk-aware semantic prediction and depth-derived training signals to estimate a continuous traversability score.
Vision--language approaches use a VLM more directly, asking it for a score or prompting it for terrain classes and weighting them with predefined values~\cite{anytraverse,zest,catnav}.
TravPro keeps the VLM frozen, does not prompt it, and fits a preference readout on its patch tokens from ordered pairs.
Knowledge distillation transfers representations from large teachers to efficient dense models.
AM-RADIO consolidates several vision foundation models into one~\cite{amradio}, and Theia distills diverse visual teachers for robot learning~\cite{theia}.
Prototype-based distillation transfers class-centered feature relations from segmentation teachers to dense students~\cite{ifvd,cirkd}.
TravPro uses fixed VLM prototypes differently.
They serve as appearance references for a preference readout rather than as class representatives, so two regions with the same semantic label can receive different scores.

\begin{figure*}[t]
\centering
\includegraphics[width=0.9\textwidth]{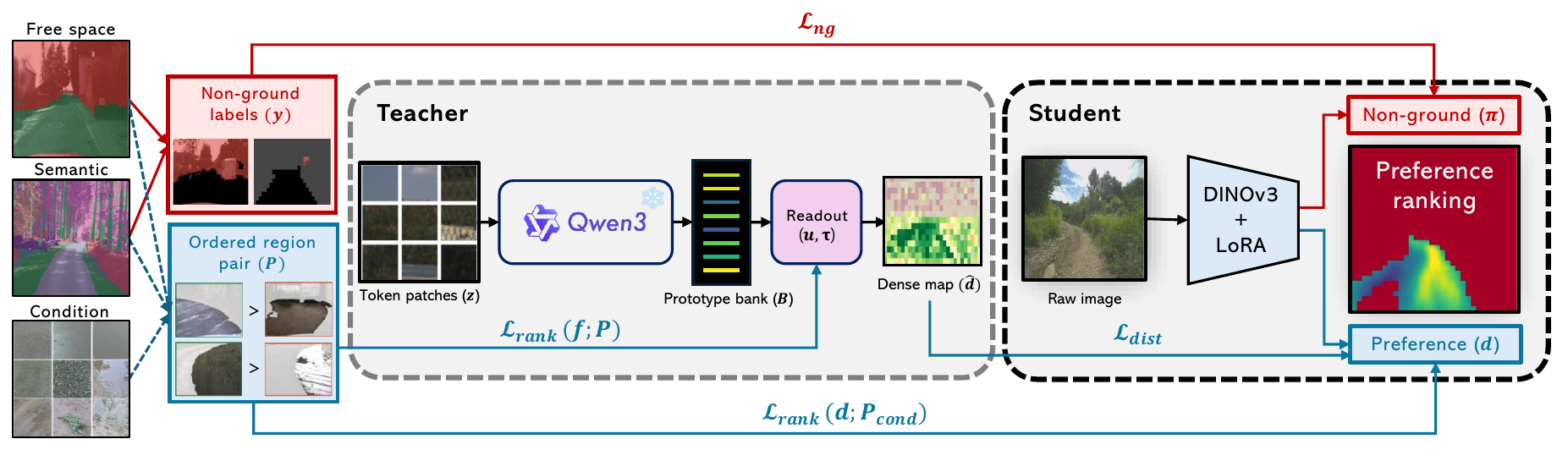}
\caption{TravPro overview. Annotations yield dense non-ground labels and ordered region pairs $\mathcal{P}$ (preferred region on the left). A frozen VLM's patch tokens are clustered once into a prototype bank; a readout fit from the pairs produces a dense teacher map $\hat d$. A student with LoRA distills $\hat d$ ($\mathcal{L}_{\mathrm{dist}}$), receives the condition pairs ($\mathcal{L}_{\mathrm{rank}}$), learns non-ground ($\mathcal{L}_{\mathrm{ng}}$), and outputs $d$ and $\pi$.}
\label{fig:overview}
\vspace{-0.3cm}
\end{figure*}

\section{METHODOLOGY}
\label{sec:method}

\subsection{Problem Formulation and Overview}
\label{sec:problem}
Given an RGB image $I\in\mathbb{R}^{H\times W\times 3}$, the deployed student $\mathcal{S}_{\theta}$ predicts a non-ground probability map $\pi$ and a terrain-preference map $d$:
\begin{equation}
(\pi,d)=\mathcal{S}_{\theta}(I).
\label{eq:student-output}
\end{equation}
Here, $\pi\in[0,1]^{h\times w}$ estimates whether a location is non-ground, such as sky, vehicles, and structures, and therefore excluded from terrain ranking, and $d\in\mathbb{R}^{h\times w}$ assigns a relative preference to the locations that remain, both at a common resolution $h\times w$.
Mud and puddles are ground and are ranked, receiving low preference through $d$.
We use $d$ only through relative order, not as an absolute measure of traversability.
At deployment, $\mathcal{S}_{\theta}$ requires only RGB input, without region masks or VLM inference.

The two outputs are supervised differently because the available labels are asymmetric.
Non-ground regions are annotated densely, whereas preference is not annotated at any resolution.
TravPro therefore recovers ordering information from the available annotations as ordered region pairs, fits a preference readout on a fixed bank of frozen VLM prototypes from these pairs, and applies it to every patch to obtain a dense teacher map.
An RGB student distills this map, imposes the condition pairs on its own output, and learns the non-ground output from the dense labels.
A lightweight residual head optionally adapts the ranking to a target domain from a few pairs.

\subsection{Pairwise Preference Supervision}
\label{sec:pairs}
The two outputs of $\mathcal{S}_{\theta}$ are supervised by different annotations.
Dense non-ground labels come from merging the background classes of freespace and semantic annotations (sky, vehicles, structures, and vegetation that is not ground cover) into one label.
These pixel-wise labels supervise $\pi$ in Sec.~\ref{sec:student}. The patch tokens whose locations carry no non-ground label, which we call ground tokens, build the prototype bank in Sec.~\ref{sec:teacher}.

Preference has no dense labels.
TravPro instead converts the ordering information that the annotations already imply into an ordered-pair set
\begin{equation}
\mathcal{P}=\mathcal{P}_{\mathrm{sem}}\cup\mathcal{P}_{\mathrm{cond}}.
\label{eq:pairs}
\end{equation}
Each pair $(R^{+},R^{-})\in\mathcal{P}$ asserts that region $R^{+}$ should be preferred to region $R^{-}$.
A region is defined by a pixel-level mask, and every loss in this paper compares two regions through the mean score of the patch tokens their masks cover, so pairs are formed at the region level, not per pixel or per token.
The two sources define regions differently.
For $\mathcal{P}_{\mathrm{sem}}$, a region is the mask of one class in one off-road image, taken from a freespace or semantic annotation.
We fix a small set of class relations whose order is clear, such as asphalt over gravel, safe road over its boundary zone, or soil over puddle, and for each relation we pair a region of the preferred class against a region of the other, possibly from a different image.
Labels such as puddle and water are annotated inconsistently, marked in some images and absorbed into the surrounding class in others, so a dense model receives contradictory targets for the same appearance. A pair uses only regions that are marked, placing them on the lower side against dry ground of the same material, as in soil over puddle, so the ordering it teaches is reliable.
For $\mathcal{P}_{\mathrm{cond}}$, a region is an entire close-up road-surface image, since a dataset such as RSCD labels each image as a whole without a pixel mask. All patch tokens of the image form one region, and we pair two images whose labels stand in a clear condition relation, such as a material-matched dry--wet comparison.
All pairs are constructed automatically from existing annotations and specify only a direction of preference. $\mathcal{P}$ is a partial order, not a complete terrain hierarchy.

\subsection{Frozen VLM Prototype Preference Teacher}
\label{sec:teacher}
Let $\Phi_{\ell}$ denote the visual encoder of a frozen VLM, Qwen3-VL 4B~\cite{qwen3vl} in our case, truncated at layer $\ell$.
For an image $I$, a single forward pass without any prompt yields the patch-token grid
\begin{equation}
\Phi_{\ell}(I)=[z_{ij}]_{i=1,j=1}^{h,w}\in\mathbb{R}^{h\times w\times C},
\label{eq:tokens}
\end{equation}
where $z_{ij}\in\mathbb{R}^{C}$ is the token at grid location $(i,j)$; for our backbone, $h\times w=15\times 28$.
We use an intermediate layer ($\ell=18$); Sec.~\ref{sec:design} compares layers.
Every token is standardized with a mean and standard deviation computed once over the full training corpus and then $\ell_{2}$-normalized; no per-image statistics are used.
The ground tokens defined in Sec.~\ref{sec:pairs}, pooled from off-road scenes and close-up road-surface images, are clustered by $k$-means into a prototype bank $B\in\mathbb{R}^{K\times C}$ ($K=256$) that is kept fixed thereafter.
Because the bank is built from appearance alone, surfaces of the same material in different conditions are not merged into one prototype, although the unsupervised clustering does not separate them perfectly.

A token $z$ is read through its soft assignment over the bank, a distribution over which prototype $z$ matches,
\begin{equation}
\alpha_{k}(z)=\operatorname{softmax}_{k}\!\big(z^{\top}B_{k}/\tau\big),\quad
f(z)=\sum_{k=1}^{K}\alpha_{k}(z)\,u_{k},
\label{eq:readout}
\end{equation}
where $f(z)$ is the readout score of the token, $u\in\mathbb{R}^{K}$ holds one preference score per prototype, and $\tau$ is a temperature.
For a score function $s$ and a pair set $\mathcal{Q}$, we use the logistic pairwise ranking loss~\cite{ranknet}
\begin{equation}
\mathcal{L}_{\mathrm{rank}}(s;\mathcal{Q})=-\sum_{(R^{+},R^{-})\in\mathcal{Q}}\log\sigma\!\big(\bar{s}(R^{+})-\bar{s}(R^{-})\big),
\label{eq:rank-loss}
\end{equation}
where $\sigma$ is the sigmoid, $\bar{s}(R)$ is the mean of $s$ over the tokens of $R$, and $\bar{f}(R)$ and $\bar{\alpha}_{k}(R)$ are the means of $f$ and $\alpha_{k}$ over the tokens of $R$.
The teacher minimizes $\mathcal{L}_{\mathrm{rank}}(f;\mathcal{P})$ over $u$ and $\log\tau$ while the VLM, the normalization statistics, and $B$ stay fixed.
Two properties follow from Eq.~\eqref{eq:readout}.
First, since $\sum_{k}\alpha_{k}=1$, adding a constant to $u$ leaves every pair unchanged, so the pairs determine $u$ only up to a shift and the gradient is zero along that direction. The scale of $u$ is bounded by the loss margin and weight decay, and it does not affect any comparison.
Second, because $\bar{f}$ is linear in $u$, the gradient of a pair on $u_{k}$ is proportional to $\bar{\alpha}_{k}(R^{+})-\bar{\alpha}_{k}(R^{-})$, the difference in average assignment weight.
Prototypes that both regions use in equal proportion receive no gradient on their scores, so each pair updates only the scores of the prototypes that one side uses more than the other.

After training, the same readout is applied to every patch of every image,
\begin{equation}
\hat{d}(I)[i,j]=\big(f(z_{ij})-m\big)/\sigma_f,
\label{eq:teacher-map}
\end{equation}
where $m$ and $\sigma_f$ are the mean and standard deviation of $f$ over the ground tokens of the training set.
Since any shift or scale of $f$ preserves the order, this standardization fixes one target for the student to regress.
Because the token normalization statistics, $B$, $u$, $\tau$, and $(m,\sigma_f)$ are all global constants, a patch receives the same value regardless of the image it appears in, and $\hat{d}$ is comparable across images and domains without per-image normalization.

\subsection{Dense RGB Student}
\label{sec:student}
The student $\mathcal{S}_{\theta}$ consists of a frozen DINOv3 backbone~\cite{dinov3} with low-rank adaptation (LoRA)~\cite{lora}, a lightweight projection to $C'$-dimensional features $\phi_{p}$ at each grid location $p$, and two heads that read $\pi$ and $d$ from $\phi_{p}$.
No decoder is used, since the teacher target exists only at the token grid; $\pi$ is predicted at the same grid so that the two heads share one resolution and $d$ can be masked by $\pi$ directly.
The student is trained with
\begin{equation}
\mathcal{L}_{\mathrm{stu}}=\mathcal{L}_{\mathrm{ng}}+w_{d}\,\mathcal{L}_{\mathrm{dist}}+w_{r}\,\mathcal{L}_{\mathrm{rank}}(d;\mathcal{P}_{\mathrm{cond}}),
\label{eq:student-loss}
\end{equation}
whose terms correspond to the non-ground labels, the teacher map, and the condition pairs, with weights $w_{d}$ and $w_{r}$.

$\mathcal{L}_{\mathrm{ng}}=\big\langle \mathrm{BCE}(\pi_{p},y_{p})\big\rangle_{p}$, the binary cross-entropy averaged over grid locations $\langle\cdot\rangle_{p}$, trains $\pi$ from the non-ground labels $y\in\{0,1\}^{h\times w}$, the pixel masks pooled to the grid by majority vote, with $y=1$ marking non-ground.
$\mathcal{L}_{\mathrm{dist}}=\big\langle \operatorname{smooth}\text{-}\ell_{1}(d_{p}-\hat{d}_{p})\big\rangle_{p:\,y_{p}=0}$ distills the teacher map at ground locations; for close-up road-surface images, which are entirely ground, this is the whole grid.
$\mathcal{L}_{\mathrm{rank}}(d;\mathcal{P}_{\mathrm{cond}})$ imposes the image-level condition pairs directly on the student output.
Distillation matches the teacher's standardized values, which preserves the large gaps between terrain categories but can invert the small gaps between conditions of one material.
This term therefore constrains $d$ itself on the condition ordering; the category pairs need no such term.
The two maps are kept separate at deployment, and $d$ is masked by $\pi$ when a single visualization is needed.

\subsection{Optional Target-Domain Adaptation}
\label{sec:adapt}
The deployed model is complete after Sec.~\ref{sec:student}, and its zero-shot ranking reflects the source domain.
In the field, terrain preference varies with the site, the platform, and the driving policy, so a small set of ordered pairs $\mathcal{P}_{\mathrm{tgt}}$ from a new site can shift the ranking toward that site's preference without updating the student.
We add a linear head on the same features $\phi_{p}$ from which $d$ is read (Sec.~\ref{sec:student}) and use it as a residual on $d$,
\begin{equation}
d'(p)=d(p)+w^{\top}\phi_{p}+b,\qquad w\in\mathbb{R}^{C'},\ b\in\mathbb{R},
\label{eq:residual}
\end{equation}
initialized at zero so that $d'=d$ before adaptation.
The residual is fit by minimizing
\begin{equation}
\mathcal{L}_{\mathrm{ada}}=\mathcal{L}_{\mathrm{rank}}(d';\mathcal{P}_{\mathrm{tgt}})+\beta\,\big\langle (w^{\top}\phi_{p}+b)^{2}\big\rangle_{p}.
\label{eq:adapt-loss}
\end{equation}
The first term imposes the target pairs on their regions.
The second term, weighted by $\beta$ and averaged over all locations of the target frames, keeps $d'$ close to $d$ at locations outside those regions.
A single direction in feature space is sufficient because a site change alters which appearances are preferred rather than how they are represented, and a few pairs cannot constrain a richer correction.
Since the optimization starts at the deployed $d$ and is stopped by a small validation split of the site, the residual stays small when the pairs are few, and one distilled model serves every site with $C'+1$ site-specific parameters.

\section{EXPERIMENTS}
\label{sec:exp}

\subsection{Setup}
\label{sec:setup}
\textbf{Training data.}
The teacher and student are trained on GOOSE~\cite{goose}, ORAD~\cite{orad}, and the even-indexed half of RSCD~\cite{rscd}.
GOOSE and ORAD supply the non-ground labels, the ground tokens, and $\mathcal{P}_{\mathrm{sem}}$; RSCD supplies $\mathcal{P}_{\mathrm{cond}}$.
In total the teacher is fit on 22{,}800 pairs: 7{,}500 region pairs from the freespace classes of ORAD, 8{,}400 region pairs from seven GOOSE class relations, and 6{,}900 image pairs from RSCD, 46 class pairs of the kinds listed in Table~\ref{tab:bench}, each sampled 150 times.
The student receives the same RSCD pairs directly.
No evaluation frame is used at any training stage, including prototype construction.

\textbf{Evaluation.}
We report pairwise ranking accuracy, the fraction of ordered pairs $(R^{+},R^{-})$ for which the mean of $d$ over $R^{+}$ exceeds that over $R^{-}$; $\pi$ is evaluated separately by pixel AUROC against non-ground masks derived from semantic labels.
Table~\ref{tab:bench} lists the six benchmarks. Five domains are unseen and form Table~\ref{tab:main}; RSCD is reported separately in Table~\ref{tab:kinds}.
\begin{table}[t]
\centering
\footnotesize
\setlength{\tabcolsep}{3pt}
\caption{Evaluation benchmarks. Rel.: number of relations; pairs are sampled from them on semantic-mask regions, or on whole images for RSCD. Trail-Wet and WayFAST pairs are human-annotated. RSCD is a training domain; its held-out frames are disjoint from training. $A\succ B$: $A$ is preferred to $B$; a list on either side stands for each combination.}
\label{tab:bench}
\begin{tabularx}{\columnwidth}{@{}l r r >{\raggedright\arraybackslash}X@{}}
\toprule
Benchmark & Rel. & Pairs & Relations \\
\midrule
Trail-Wet (ours) & -- & 204 & human preference on a rainy mountain trail recorded by our UGV; three annotators agree on 202 \\
ACDC~\cite{acdc} & 2 & 8{,}000 & urban road: dry $\succ$ wet, dry $\succ$ snow-covered \\
Offroad-Sem (ours) & 8 & 16{,}000 & campus and mountain trail recorded by our UGV: grass $\succ$ high grass; soil, dirt road, asphalt $\succ$ gravel; dirt road $\succ$ high grass; soil, dirt road, grass $\succ$ puddle \\
RELLIS~\cite{rellis} & 6 & 12{,}000 & off-road: concrete, grass $\succ$ mud, puddle, rubble \\
RSCD~\cite{rscd} (held-out frames) & 10 & 20{,}000 & close-up road surface, within one material. Cover: bare asphalt $\succ$ ice, fresh snow; dry, wet $\succ$ water-covered. Same material: dry $\succ$ wet gravel, mud; smooth $\succ$ severe asphalt, concrete \\
WayFAST~\cite{wrizz} & -- & 1{,}738 & farm and trail: W-RIZZ's own point pairs \\
\bottomrule
\vspace{-0.4cm}
\end{tabularx}
\end{table}
The evaluation relations of the five unseen domains are used in no training stage of any method, and TravPro's training pairs come only from GOOSE, ORAD, and RSCD.
Scoring by relations does not favor TravPro: a semantic lookup encodes the same category orders through its per-class values, and Sec.~\ref{sec:main} shows that it matches TravPro where the relations are between categories.
The gap appears on condition relations, which a per-class value cannot express.

\textbf{Implementation.}
Qwen3-VL 4B receives each image at $480\times 896$, which gives a $15\times 28$ token grid at layer 18; the student receives $512\times 512$ and projects to the same grid.
The teacher readout ($K=256$, $u$ and $\log\tau$) is fit with AdamW (learning rate $3\times 10^{-2}$) for 2{,}000 iterations of 48 pairs.
The student uses DINOv3-B/16 with LoRA (rank 8, on the query, key, value, and output projections) and is trained for 12{,}000 steps with AdamW, a batch of four, and learning rates of $10^{-4}$ for LoRA and $3\times 10^{-4}$ for the projection and heads under cosine annealing, with $w_{d}=w_{r}=3$ and $\beta=0.1$.
At deployment the student runs at 12\,Hz on an NVIDIA Jetson AGX Orin in PyTorch with FP16.
All numbers are means over three seeds.

\begin{table}[t]
\centering
\caption{Pairwise ranking accuracy on unseen domains, scored by $d$ alone. Best per column in bold, second best underlined. Mean is over all five benchmarks. D3-L and D3-F: DINOv3 segmenter with LoRA and with frozen weights. TW: Trail-Wet, AC: ACDC, OS: Offroad-Sem, RL: RELLIS, WF: WayFAST. Rows without a citation are variants we implemented.}
\label{tab:main}
\footnotesize
\setlength{\tabcolsep}{4pt}
\begin{tabular*}{\columnwidth}{@{\extracolsep{\fill}\hspace{\tabcolsep}} l ccccc c}
\toprule
Method & TW & AC & OS & RL & WF & Mean \\
\midrule
SemLk (D3-L)                & .799 & .625 & \textbf{.880} & \underline{.775} & .838 & \underline{.783} \\
SemLk (D3-F)                & \underline{.833} & .459 & .851 & .762 & .866 & .754 \\
SemLk +RSCD (D3-L)          & .797 & .543 & .840 & .672 & .765 & .723 \\
SemLk +RSCD (D3-F)          & .793 & .548 & .842 & .696 & .740 & .724 \\
SemLk (SAM2)                & .719 & .652 & .582 & .596 & .808 & .671 \\
Bin-confidence              & .711 & .433 & .529 & .387 & .755 & .563 \\
ViTA~\cite{vita}                        & .755 & .884 & .627 & .579 & .860 & .741 \\
Sem-VLM-Fixed               & .673 & .778 & .632 & .532 & .867 & .696 \\
AnyTraverse~\cite{anytraverse}                 & .456 & \underline{.923} & .676 & .622 & .781 & .692 \\
VLM direct-prompt             & .760 & .482 & .573 & .329 & .759 & .581 \\
W-RIZZ~\cite{wrizz}                      & .592 & .438 & .546 & .628 & \textbf{.942} & .629 \\
STEPP~\cite{stepp}                       & .430 & .419 & .653 & .413 & .846 & .552 \\
\midrule
TravPro (ours)              & \textbf{.887} & \textbf{.973} & \underline{.878} & \textbf{.942} & \underline{.897} & \textbf{.915} \\
\bottomrule
\end{tabular*}
\vspace{-0.4cm}
\end{table}

\textbf{Baselines.}
The baselines cover the three ways of Sec.~\ref{sec:intro} and two control baselines that share a component with TravPro: a semantic lookup given our RSCD condition labels as dense targets, and our teacher VLM prompted directly.
A semantic lookup (SemLk) scores each region by a fixed per-category value averaged over a segmenter's prediction, the first way; we vary the segmenter (LoRA-adapted and frozen DINOv3, and SAM2) to show that the bottleneck is the fixed value, and train two more on RSCD as dense targets (+RSCD) so that the lookup receives the same condition supervision that TravPro uses as pairs.
Two baselines derive a score from binary freespace labels alone: a binary confidence baseline uses the freespace confidence of the DINOv3 segmenter, and ViTA~\cite{vita} adapts SAM2 with pseudo-depth to regress a continuous traversability score.
AnyTraverse~\cite{anytraverse} prompts an open-vocabulary segmenter for terrain classes and attaches a predefined value to each. Sem-VLM-Fixed is our upgraded version of AnyTraverse with twelve prompts and graded weights, so that vocabulary is not the bottleneck.
VLM direct-prompt prompts our teacher Qwen3-VL with two outlined regions and asks which is preferable, testing whether prompting alone extracts the order that the readout learns.
ZeST~\cite{zest} and CATNAV~\cite{catnav} have no public code; the two VLM baselines above represent the same family.
W-RIZZ~\cite{wrizz} and STEPP~\cite{stepp} are trained from sparse pairs and from a robot's own traversals, the second and third ways; we evaluate the authors' released models, since both need data only their platforms provide.

\subsection{Main Results}
\label{sec:main}
TravPro reaches a mean accuracy of 0.915 on the five unseen domains against 0.783 for the strongest baseline, a semantic lookup on a LoRA-adapted segmenter (Table~\ref{tab:main}).
The size of the gap depends on what kind of relation a benchmark scores, category or condition.
On Offroad-Sem, whose relations are mostly between categories, the lookup matches TravPro, since a category order is exactly what a per-class value encodes.
On Trail-Wet, ACDC, and RELLIS, whose relations turn on surface condition, the best baseline falls behind by between 5 and 17 points.
On RELLIS the lookup fails for two reasons.
Its segmenter reads mud as soil, and puddle labels are too inconsistent to train a dense target (Sec.~\ref{sec:pairs}).

\begin{figure}[t]
\centering
\includegraphics[width=\columnwidth]{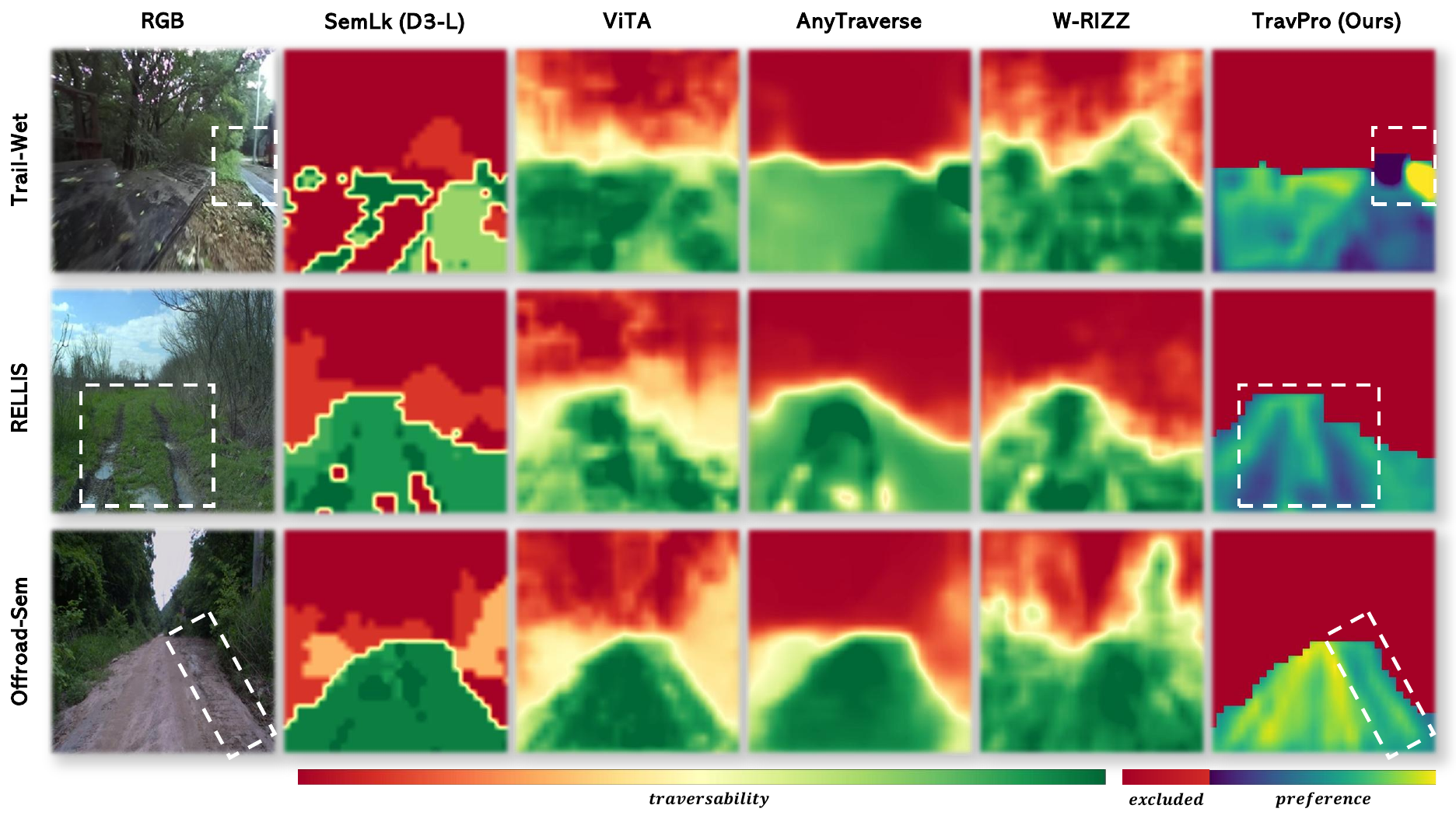}
\caption{Qualitative comparison on unseen domains. Baselines output a traversability score (red to green); TravPro outputs terrain preference (dark to yellow), upsampled bilinearly from the $15\times 28$ grid for display, with non-ground in red at grid resolution. Dashed boxes mark the regions discussed in Sec.~\ref{sec:qual}.}
\label{fig:qual}
\vspace{-0.4cm}
\end{figure}

Two comparisons in Table~\ref{tab:main} hold a component of TravPro fixed and change only how it is used.
Prompting the teacher VLM itself to compare two regions gives 0.581, below chance on RELLIS.
Reading the same model's tokens through the learned readout gives 0.915.
Giving the semantic lookup the RSCD labels as dense targets (+RSCD) lowers its unseen mean, whereas TravPro learns its condition axis from the same labels converted into ordered pairs.
On W-RIZZ's own WayFAST pairs, TravPro is second only to W-RIZZ.
The non-ground head transfers without adjustment, with a mean pixel AUROC of 0.958 over the five unseen domains, and Fig.~\ref{fig:qual} shows it excluding vehicles, trees, and sky.

\begin{table}[t]
\centering
\caption{Accuracy on the ten RSCD relations of Table~\ref{tab:bench}, split into cover, where another material lies on the surface, and same material, where the surface itself changes. Worst: lowest single-relation accuracy. Best in bold, second best underlined.}
\label{tab:kinds}
\footnotesize
\setlength{\tabcolsep}{3pt}
\begin{tabular*}{\columnwidth}{@{\extracolsep{\fill}\hspace{\tabcolsep}} l cc cc cc}
\toprule
\multirow{2}{*}{Method} & \multicolumn{2}{c}{Cover} & \multicolumn{2}{c}{Same material} & \multicolumn{2}{c}{All (10)} \\
\cmidrule(lr){2-3}\cmidrule(lr){4-5}\cmidrule(lr){6-7}
& Snow/ice & Water & Wet & Rough & Mean & Worst \\
\midrule
SemLk (D3-L)             & .776 & .769 & .391 & .504 & .642 & .310 \\
SemLk (D3-F)             & .861 & .787 & .565 & .674 & .735 & .492 \\
SemLk +RSCD (D3-L)       & \textbf{.999} & \underline{.921} & \textbf{.905} & \underline{.859} & \underline{.921} & \underline{.814} \\
SemLk +RSCD (D3-F)       & \underline{.997} & .887 & \underline{.871} & .838 & .896 & .750 \\
SemLk (SAM2)             & .708 & .689 & .452 & .699 & .647 & .384 \\
Bin-confidence           & .536 & .686 & .424 & .791 & .625 & .310 \\
ViTA~\cite{vita}                     & .815 & .591 & .367 & .674 & .607 & .286 \\
Sem-VLM-Fixed            & .851 & .811 & .291 & .641 & .681 & .237 \\
AnyTraverse~\cite{anytraverse}              & .704 & .843 & .387 & .598 & .675 & .364 \\
VLM direct-prompt          & .645 & .696 & .518 & .843 & .680 & .403 \\
W-RIZZ~\cite{wrizz}                   & .826 & .598 & .348 & .485 & .571 & .325 \\
STEPP~\cite{stepp}                    & .785 & .557 & .357 & .319 & .515 & .248 \\
\midrule
TravPro (ours)           & .885 & \textbf{.963} & \textbf{.905} & \textbf{.973} & \textbf{.938} & \textbf{.837} \\
\bottomrule
\end{tabular*}
\vspace{-0.4cm}
\end{table}

\subsection{Where the Condition Axis Comes From}
\label{sec:condition}
A semantic taxonomy expresses surface condition only where it draws a class boundary, and Table~\ref{tab:kinds} shows this on RSCD.
Where another material covers the surface, as with snow or standing water, the taxonomy has a class and a semantic lookup is competitive.
Where only the condition changes, as with wet gravel or rough asphalt, it has no class, and the lookup falls to 0.565 and 0.674 against 0.905 and 0.973 for TravPro.
Training the lookup on RSCD labels as dense targets adds those boundaries and recovers RSCD itself, but the gain stays there: both +RSCD lookups fall on the unseen mean (Table~\ref{tab:main}), whereas the same labels as ordered pairs raise TravPro on every unseen domain.
Whether the condition axis transfers is decided by how the supervision is used, not by where it comes from.

\begin{figure}[t]
\centering
\includegraphics[width=0.9\columnwidth]{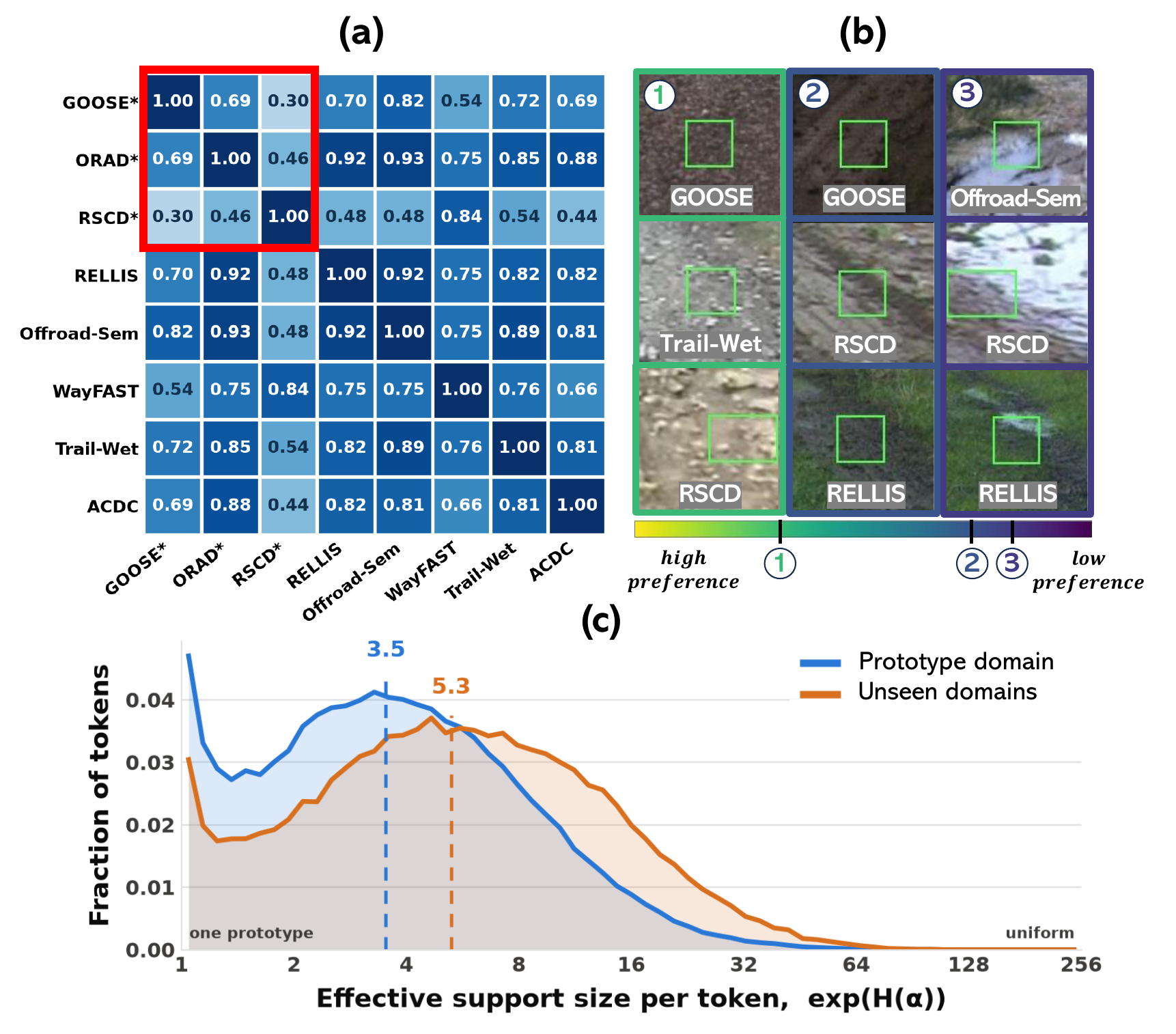}
\caption{(a) Bhattacharyya coefficient between the prototype-usage distributions, the mean soft assignment of a domain's tokens, of the three prototype domains (red box), the corpora that built the bank, and five unseen domains. (b) Patches grouped by assigned prototype, with columns ordered by learned preference and marked on the bar; GOOSE and RSCD patches are from held-out frames, the others from unseen domains. (c) Effective number of prototypes per token for prototype domains and unseen domains; dashed lines are medians.}
\vspace{-0.4cm}
\label{fig:bank}
\end{figure}

Table~\ref{tab:ablation} shows where TravPro obtains this axis.
Adding the condition pairs directly to the student as the third term of Eq.~\eqref{eq:student-loss} raises every benchmark over distillation alone.
The gain grows with how much the benchmark concerns condition, from 3.6 points on Offroad-Sem to 14.1 on RELLIS.
Distillation alone carries the teacher's values but loses part of the order between conditions, and the direct term restores it.
Removing RSCD from teacher and student lowers accuracy on RSCD itself to 0.715.
Trail-Wet and ACDC, however, still score above every semantic lookup, so the condition axis does not depend on RSCD alone.
The off-road pairs already carry the axis, as Sec.~\ref{sec:intro} argued from the puddle and water labels, and RSCD extends the axis further.
Fig.~\ref{fig:bank}a shows how.
The close-up view of RSCD overlaps the off-road prototype domains only partly, at 0.30 and 0.46, but the prototypes it does share are where its condition relations reach off-road terrain.
Every unseen domain in turn overlaps at least one prototype domain at 0.84 or above, so unseen tokens are read through prototypes the pairs have already ordered, which is indirect evidence that the ordering generalizes.
On unseen domains a token also spreads over more prototypes than on the prototype domains (Fig.~\ref{fig:bank}c), so the order reaches appearances between them.

\subsection{Qualitative Results}
\label{sec:qual}
Fig.~\ref{fig:bank}(b) groups held-out and unseen patches by assigned prototype, ordered by learned preference.
Patches of one appearance gather on one prototype regardless of domain and viewpoint, and increasingly wet ground moves to prototypes of lower rank, from dry gravel to rutted wet soil to standing water.
Fig.~\ref{fig:qual} compares $d$ masked by $\pi$ with the baselines.
The baselines assign one flat value to the drivable area, whereas TravPro varies within it, ranking the paved sidewalk above the wet trail, puddles and mud below the grass around them, and mud in tire ruts below the dry soil beside them.

\begin{table}[t]
\centering
\footnotesize
\setlength{\tabcolsep}{3pt}
\caption{Ablations. Each row changes one element of the full model, which uses Qwen3-VL 4B layer 18 tokens, a 257-parameter prototype readout, and a DINOv3-B/16 student with LoRA. Both: RSCD removed from teacher and student; FT: full fine-tuning. RS: RSCD; other columns as in Table~\ref{tab:main}.}
\label{tab:ablation}
\begin{tabular*}{\columnwidth}{@{\extracolsep{\fill}\hspace{\tabcolsep}}lccccccc@{}}
\toprule
 & TW & RS & AC & OS & RL & WF & Mean \\
\midrule
TravPro (full) & .887 & .938 & .973 & .878 & .942 & .897 & .919 \\
\midrule
\multicolumn{8}{l}{\emph{Pair supervision}} \\
\; w/o student $\mathcal{L}_{\mathrm{rank}}$ & .834 & .826 & .851 & .842 & .801 & .861 & .836 \\
\; w/o RSCD (both)         & .841 & .715 & .847 & .814 & .841 & .868 & .821 \\
\midrule
\multicolumn{8}{l}{\emph{Teacher tokens}} \\
\; DINOv3-B/16             & .804 & .872 & .889 & .732 & .851 & .842 & .831 \\
\; Qwen3-VL, layer 5       & .873 & .927 & .936 & .903 & .938 & .875 & .909 \\
\; Qwen3-VL, layer 27      & .799 & .930 & .971 & .871 & .933 & .858 & .894 \\
\midrule
\multicolumn{8}{l}{\emph{Teacher readout}} \\
\; MLP (656K)              & .819 & .927 & .977 & .917 & .938 & .902 & .913 \\
\midrule
\multicolumn{8}{l}{\emph{Student encoder}} \\
\; DINOv3, frozen          & .824 & .914 & .938 & .878 & .934 & .873 & .893 \\
\; DINOv3, full FT         & .863 & .939 & .954 & .857 & .932 & .872 & .902 \\
\; RADIO, LoRA             & .873 & .924 & .794 & .863 & .936 & .868 & .876 \\
\; SAM2, LoRA              & .632 & .852 & .597 & .752 & .918 & .873 & .770 \\
\bottomrule
\end{tabular*}
\end{table}

\subsection{Design Choices}
\label{sec:design}
The remaining rows of Table~\ref{tab:ablation} isolate the other components.
Replacing the VLM tokens with DINOv3 tokens lowers every benchmark, most on Offroad-Sem and RELLIS, so the VLM tokens carry a separation of terrain categories and conditions that DINOv3 tokens lack, as Sec.~\ref{sec:intro} hypothesized.
Among the VLM's layers, an intermediate one (18) transfers best.
Replacing the prototype readout with a two-layer MLP on the same tokens and pairs gives 0.913 against 0.919, although the MLP has 656K parameters and the prototype readout 257.
The gap is largest on Trail-Wet, the human-judged benchmark on surface condition, where the prototype readout leads by 6.8 points.
A pair updates only the prototypes that differ between its two regions (Sec.~\ref{sec:teacher}), so the direction of each training relation is stored in those prototypes and reaches every region that uses them, including appearances no relation names.
We therefore keep the readout as one score per prototype and one temperature instead of an MLP.
For the student, LoRA on DINOv3 beats full fine-tuning and a frozen backbone.
The number of prototypes, from 64 to 1{,}024, and the loss weights change the mean by less than one point.

In the teacher, the soft assignment $\alpha$ does not collapse to one prototype per token, as the readout of Sec.~\ref{sec:teacher} assumes.
For each token we count how many prototypes are needed to reach 90\% of its assignment weight $\alpha$; the count is seven on average (median four) out of 256 (Fig.~\ref{fig:bank}c), so a token's score is a weighted average of several prototype scores and changes smoothly across a terrain type instead of taking one fixed value as in a lookup.

\begin{figure}[t]
\centering
\includegraphics[width=\columnwidth]{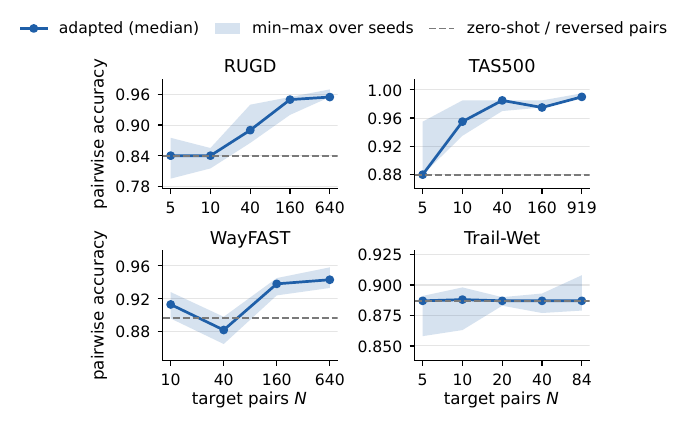}
\caption{Target-domain adaptation with $N$ ordered pairs on four unseen domains. Evaluation pairs are disjoint from the adaptation pairs at the image level; the largest $N$ is the full pool of each domain.}
\label{fig:adapt}
\vspace{-0.4cm}
\end{figure}

\subsection{Target-Domain Adaptation}
\label{sec:adapt-exp}
Fig.~\ref{fig:adapt} evaluates the residual head of Sec.~\ref{sec:adapt} offline on four unseen domains: Trail-Wet and WayFAST from Table~\ref{tab:bench}, and RUGD~\cite{rugd} and TAS500~\cite{tas500}, with relations derived from their semantic labels as for Offroad-Sem.
Adaptation pairs are built the same way as each domain's evaluation pairs, but from frames disjoint from the evaluation frames.
A small validation split of the target domain receives no gradient.
It only selects which step's weights are kept, and if no step improves on the validation split, it keeps step zero, $w=0$.
On RUGD and TAS500, whose relations are mostly between categories, the gain is large and arrives early: ten TAS500 pairs raise accuracy from 0.880 to 0.955, and 160 RUGD pairs raise it from 0.840 to 0.950.
On Trail-Wet, which measures the condition axis, the median stays at the zero-shot value for every $N$.
With only 84 adaptation pairs, the validation split is too small to distinguish an improving step from noise, and the selection falls back to step zero.
WayFAST falls slightly below zero-shot at $N=40$, within the range across seeds, and rises above it from $N=160$ on, reaching 0.943 at $N=640$.
If the weights are taken from the last step instead of being selected on the validation split, RUGD and TAS500 gain almost as much, but reversing every target pair collapses accuracy by between 65 and 83 points.
Reversing every target pair, to test whether the split protects against wrong pairs, leaves accuracy at the zero-shot value in every domain and for every $N$.
If the weights are taken from the last step instead, correct pairs still raise RUGD and TAS500 almost as much, but the reversed pairs collapse accuracy by between 65 and 83 points.
The gain thus comes from the pairs, and the split prevents damage when they are wrong.

\section{CONCLUSIONS}
We presented TravPro, which learns a dense terrain-preference ranking by reformulating existing freespace, semantic, and road-surface annotations as ordered region pairs.
Non-ground regions are learned directly from dense labels.
The pairs fit a 257-parameter readout on a fixed bank of frozen VLM prototypes, and a student distills the result and outputs both fields from RGB alone.
On five unseen domains the student transfers a condition-sensitive ordering that a semantic lookup cannot represent and that prompting the same VLM does not produce; what transfers depends on how the supervision is used, not on where it comes from.
The prototype bank is built by unsupervised clustering and partly separates surface conditions of the same terrain.
Future work includes a readout that uses the VLM's features beyond a fixed bank, and grounding the learned visual order in robot embodiment, for instance by collecting adaptation pairs during deployment.








\bibliographystyle{ieeetr}
\bibliography{ref}

\end{document}